\documentclass[runningheads]{llncs}

\usepackage[T1]{fontenc}
\usepackage{graphicx}
\usepackage{hyperref}
\usepackage{ amssymb }
\usepackage{booktabs}
\usepackage{array}
\usepackage{amsmath}
\usepackage{dsfont}
\usepackage{adjustbox}
\usepackage{float}
\usepackage{subfiles}

\hypersetup{
    colorlinks=true,
    citecolor=blue,
    urlcolor=red,
}

\begin{document}

\title{SelectiveKD: A semi-supervised framework for cancer detection in DBT through Knowledge Distillation and Pseudo-labeling}
\titlerunning{SelectiveKD for cancer detection in DBT}

\author{Laurent Dillard\inst{1}\thanks{These authors contributed equally to this work.}, Hyeonsoo Lee\inst{1,*}, Weonsuk Lee\inst{1,*}, Tae Soo Kim\inst{1}, Ali Diba\inst{1} and Thijs Kooi\inst{1}}
%

\institute{
Lunit Inc, Seoul, Republic of Korea\\
\email{\{laurent.dillard, hslee, iwonseok5762, taesoo.kim, ali, tkooi\}@lunit.io}}

\maketitle              

\begin{abstract}
When developing Computer Aided Detection (CAD) systems for Digital Breast Tomosynthesis (DBT), the complexity arising from the volumetric nature of the modality poses significant technical challenges for obtaining large-scale accurate annotations. Without access to large-scale annotations, the resulting model may not generalize to different domains. Given the costly nature of obtaining DBT annotations, how to effectively increase the amount of data used for training DBT CAD systems remains an open challenge.

In this paper, we present SelectiveKD, a semi-supervised learning framework for building cancer detection models for DBT, which only requires a limited number of annotated slices to reach high performance. We achieve this by utilizing unlabeled slices available in a DBT stack through a knowledge distillation framework in which the teacher model provides a supervisory signal to the student model for all slices in the DBT volume. Our framework mitigates the potential noise in the supervisory signal from a sub-optimal teacher by implementing a selective dataset expansion strategy using pseudo labels. 

We evaluate our approach with a large-scale real-world dataset of over 10,000 DBT exams collected from multiple device manufacturers and locations. The resulting SelectiveKD process effectively utilizes unannotated slices from a DBT stack, leading to significantly improved cancer classification performance (AUC) and generalization performance.

\end{abstract}

\section{Introduction}
Digital Breast Tomosynthesis (DBT) offers three-dimensional imaging for breast cancer screening which improves over the traditional 2D mammography by better detecting cancer especially from dense breasts \cite{AndhariaDBT,DhamijaDBT,ko2021accuracy,Nguyendbt}, lowering false-positive recall rates \cite{bahl2018breast,Shahan2016AnOO} and demonstrating overall superior cancer detection rates \cite{chikarmane2021screening}. Because images are taken from multiple angles, lesions visibility is enhanced. However, the volumetric nature of the data can also increase the radiologists' workload caused by longer reading time as they have to go through an entire stack of images~\cite{shoshan2022artificial}. Recent advances in deep learning-based Computer-Aided Detection (CAD) systems for DBT offer a promising solution to mitigate this. 

Deep learning-based CAD systems for breast cancer detection in DBT have been an active area of research \cite{goldberg2022new,Huang2023SpectralSpatialMT,lee2023transformer,Shen2023FullyCS,tardy2021trainable}. However, existing CAD  approaches for DBT also face challenges from the complexity arising from the volumetric nature of the modality. Each DBT stack comprises roughly 40 - 80 slices for each view, resulting in hundreds of images for each exam. Moreover, slices of interest (i.e., with a suspicious lesion) comprise only a small portion of a given DBT stack. As a result, collecting large-scale accurate annotations for marking such regions of interest poses a significant technical challenge. As a compromise, practitioners often build DBT CAD systems by annotating only a sparse set of slices of interest by marking them as positive and treating all other slices as negative \cite{tamCXR,zhang2022semisupervised}. This practice limits the scale of annotated datasets for DBT and introduces potential noise in the annotations. Without accurate large-scale annotations, deep learning models over-fit easily, ultimately hindering their clinical value as CAD \cite{singh2023shifting}.

In this paper, we present a semi-supervised learning framework, \textit{SelectiveKD}, for building a cancer detection model that leverages unannotated slices in a DBT stack. Inspired by recent advancements in semi-supervised learning \cite{chen2022semi,chen2020semi,Huang2023SpectralSpatialMT,Shen2023FullyCS,xie2020self}, we build upon the concept of Knowledge Distillation (KD) \cite{hinton2015distilling} to be able to use all the unlabeled slices from a DBT exam. We first obtain a teacher model using available annotated slices, which provide a supervisory signal for the student model, which trains using both annotated and unannotated slices. 

In doing so, some noise can be introduced into the training process due to the teacher's inherent sub-optimal performance.
Our framework mitigates this using Pseudo Labels (PL) generated with the teacher \cite{berthelot2019mixmatch,lee2013pseudo} to select which unannotated slices to include in training. We empirically show that PLs correct the noise introduced in the KD process. Our framework effectively combines KD and PL to leverage all slices available in a DBT volume using only a limited number of annotated slices, leading to significantly improved cancer detection performance.

We validate our framework using a large real-world dataset of over 10,000 DBT exams, collected from multiple device manufacturers (i.e., Hologic, GE, Siemens) and institutions. We demonstrate that the presented approach not only improves detection performance but also the model's ability to generalize across different manufacturers. We show that our framework achieves this generalization capacity \textit{without} requiring annotations from the target device manufacturers. Further, by effectively using unlabeled images, we can actually reach the same level of cancer detection performance with considerably fewer labeled images, leading to large potential annotation cost savings for building practical CAD systems for DBT.
\section{Method}
\begin{figure*}[t]
    \begin{center}
    \includegraphics[width=\textwidth]{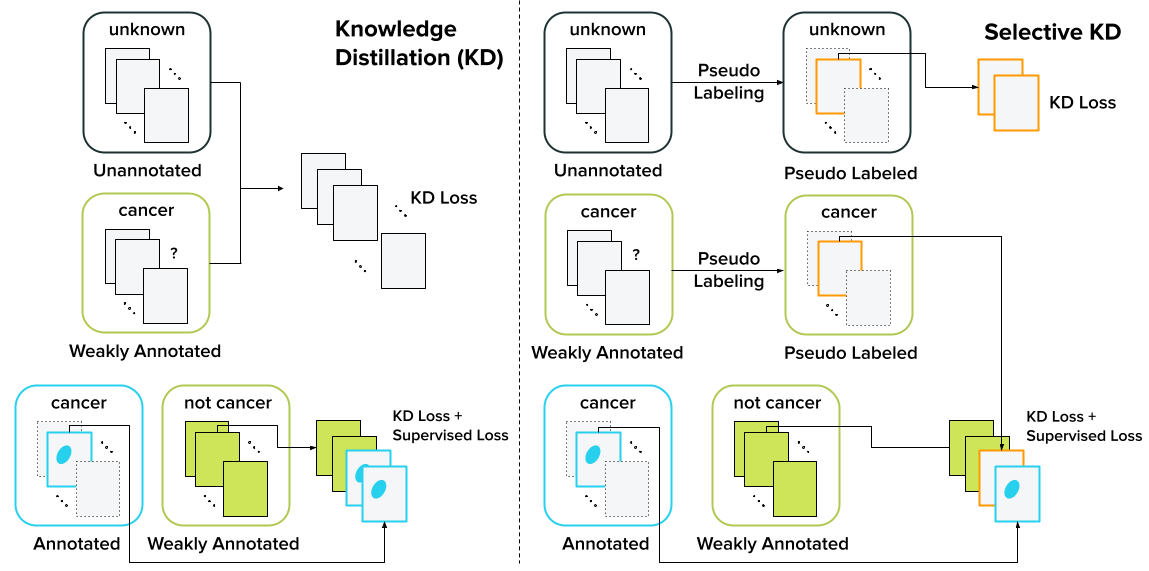}
    \vspace{-0.5cm}
    \caption{Overview of the SelectiveKD framework: Illustrating the application of knowledge distillation (KD) and SelectiveKD across varying levels of annotation of DBT slices. SelectiveKD leverages pseudo-labeling to select a few slices for training, minimizing the potential noise introduced by standard KD and reinforcing weak annotations. The approach allows use of all slices in the DBT stack during training.}
    \label{fig:method-overview}
    \end{center}
\end{figure*}

\subsection{Practical Annotation of DBT data}
A Digital breast tomosynthesis (DBT) exam contains multiple stacks of high resolution images reconstructed from X-ray images taken from different angles. The task of annotating every single image within a DBT stack, typically consisting of 40 to 80 slices, is prohibitively time-consuming and expensive. Consequently, alternative annotation strategies are used.

One strategy involves selectively annotating the slice where the lesion is most clearly visible, marking it with a contour annotation. This method reduces the annotation workload, however, a thorough examination of each slice in the DBT stack is still required to identify the most representative one for annotation. 

Although a single slice is annotated per DBT stack, it is safe to consider the lesion is roughly in the same location in adjacent slices; therefore, each stack can yield up to 3 annotated slices. In this work, annotated cases refer to cases annotated using this strategy. For annotated DBT stacks without lesions, all slices can be used for training. Still, in order to avoid a large imbalance between positive and negative slices, those stacks are sub-sampled using strided sampling. 

An alternative approach is to collect weak annotations by utilizing other diagnostic results, such as ultrasound or biopsy. This approach can only yield breast-level annotation indicating the presence or absence of cancer. These weak annotations are difficult to leverage for model training due to the variable visibility of lesions across the stack of slices. Even in DBT stacks confirmed to exhibit cancerous lesions, pinpointing the specific slices where the lesions are detectable remains elusive. Note, however, that for DBT stacks confirmed to be cancer-negative, we can safely consider all slices to be negative and use them in training the same way as for fully annotated stacks.

\subsection{Incorporating Pseudo-Labeling and Knowledge Distillation}

To address the challenge of limited annotations in DBT data for classification and segmentation tasks, our methodology leverages unannotated data through pseudo-labeling (PL) and knowledge distillation (KD), carefully navigating the inherent risk of introducing noise from automatic labeling.

\textbf{Refined Learning Through Knowledge Distillation:}  Confronting sparse annotations, our approach initially considers KD as a primary method. The model's training is refined through KD by leveraging the probabilistic outputs (soft labels) produced by the teacher model, which is pre-trained on a small annotated dataset. Unlike hard labels, soft labels provide a detailed continuum of information. This approach enriches the student model's training by incorporating the teacher's knowledge rather than just binary outcomes.

\textbf{Selective Dataset Expansion through Pseudo-Labeling Filtering:} Recognizing the potential for noise introduction inherent in direct KD, we strategically implement PL filtering to refine the dataset beforehand. This preprocessing step involves utilizing a teacher model to generate predictions for unannotated images. We only include images with predicted scores above a predefined confidence threshold as additional training data. By incorporating these high-confidence predictions, we limit the introduction of noise from incorrect labels.

We implement a dual-loss strategy, combining supervised loss and KD loss. KD loss is computed on all the selected data, while the supervised loss is only computed on data that has annotations. The combined loss, balanced by a constant parameter \(\alpha\), is designed to optimize learning from both annotated data and the nuanced insights provided by the teacher model through soft labels:

\begin{equation}
\mathcal{L} = 
\begin{cases} 
\text{for annotated or weakly-annotated data if } f(x;\theta_t) > T, \\
\quad\quad \mathcal{L}_{\text{Sup}}(y, f(\tilde{x};\theta)) + \alpha \cdot \mathcal{L}_{\text{KD}}(f(\tilde{x};\theta_t), f(\tilde{x};\theta)) \\
\text{for unannotated data if } f(x;\theta_t) > T, \\
\quad\quad \alpha \cdot \mathcal{L}_{\text{KD}}(f(\tilde{x};\theta_t), f(\tilde{x};\theta)) \\
\end{cases}
\end{equation}

where \(y\) denotes the annotations; \(f(\tilde{x};\theta)\) represents the predictions for the augmented image \(\tilde{x}\) using the student model parameters \(\theta\); \(f(\tilde{x};\theta_t)\) and \(f(x;\theta_t)\) are the soft labels from the teacher model \(\theta_t\) on the augmented and non augmented image respectively; and \(T\) represents the confidence threshold. We utilize binary cross-entropy loss for the supervised loss (\(\mathcal{L}_{\text{Sup}}\)) and mean-square error loss for the KD loss (\(\mathcal{L}_{\text{KD}}\)). This formulation is uniformly applied across both classification and segmentation tasks. Only the KD loss is applied for weakly annotated data in segmentation, as direct annotations are unavailable. This approach ensures that the model learns effectively from both direct annotations and the distilled insights of the teacher model.
\section{Experiments}
\subsection{Dataset}

Our proprietary DBT dataset encompasses a total of 13,150 four-view DBT exams from multiple U.S. institutions. The dataset includes 2,487 cancer-positive exams confirmed by biopsy, 3,398 normal exams, and 7,265 benign findings, of which 2,773 are biopsy-confirmed, and 4,492 have a follow-up period of at least one year confirming benign status. The dataset comprises 6,823 exams from recorded with a Hologic device, 2,353 exams from GE and 3,974 exams from a Siemens device. All exams recorded with Hologic devices are annotated, containing both breast level and slice level annotations. The GE and Siemens data are only weakly annotated and contain only breast level annotations.

The dataset was split into training (9,174 exams), validation (1,988 exams), and test (1,988 exams) sets. The allocation was performed, ensuring a balanced representation of cancer, benign, and normal exams across each set.

Our datasets are de-identified in strict adherence to the HIPAA Safe Harbor guidelines. This process involves the removal of all Protected Health Information (PHI) from both the image pixels and the DICOM header. Because data is de-identified Institutional Review Board (IRB) approval is not needed.

\subsection{Implementation Details}
Our model, developed on Pytorch, was distributed over eight NVIDIA V100 GPUs for training. Following the training protocol, we adopted the stochastic gradient descent (SGD) optimizer. We trained for 60,000 iterations with an initial learning rate of 0.012, weight decay of 0.0001, and momentum of 0.9, employing a cosine annealing strategy for learning rate adjustment \cite{loshchilov2016sgdr}. Image inputs were resized to 896$\times$640 pixels and processed in batches of 192. Data augmentation included geometric modifications such as horizontal and vertical flipping and rotations, along with photometric transformations like brightness and contrast adjustments, Gaussian noise, sharpening, CLAHE, and solarization. Our architecture is built upon a ResNet-34 backbone, pre-trained on ImageNet, serving as the foundation for both the teacher and student models in our semi-supervised learning setup. The confidence thresholds pseudo-labeling (PL) sampling was selected through grid search and fixed to 0.7 when using weak annotations and 0.1 without. The stride for sampling slices in cancer-negative DBT stacks is 2. For the weighting of the KD loss, we set \(\alpha_{\text{seg}} = 25\) for segmentation and \(\alpha_{\text{clf}} = 1\) for classification.

\subsection{Experimental setting}
We first trained a model using only annotated exams to serve as the teacher model for our semi-supervised experiments with knowledge distillation (KD) and provide a baseline for comparison. For the semi-supervised experiments, we considered several strategies for selecting slices from the weakly annotated datasets. The different experiment settings are as follows:
\begin{enumerate}
    \item \textbf{Baseline}: In this setting, we only use the annotated dataset for training, and KD is not used. This baseline is used as the teacher model for the semi-supervised experiments.
    \item \textbf{KD}: In this setting, we use KD and also include weakly annotated data in training but ignore weak annotations, therefore considering the added data as unannotated.
    \item \textbf{KD with weak annotations}: In this setting, we use KD and include weakly annotated data in training. For cancer-positive breasts, we select all DBT slices. Although we know the breast is cancer-positive, we don't know which exact slices contain the lesion, so only KD loss is computed.
    \item \textbf{SelectiveKD}: This setting is the same as KD, but we also include pseudo-labeling filtering.
    \item \textbf{SelectiveKD with weak annotations}:  This setting is the same as KD with weak annotations. However, we also include pseudo-labeling filtering and use supervised and KD loss on selected slices.
\end{enumerate}

\subsection{Results}

Models are compared using the Area Under the Receiver Operating Characteristic Curve (AUC). To compute confidence intervals and compare models, we utilized the DeLong test~\cite{delong1988comparing}. Results are shown in table \ref{tab:main-results}. 

\begin{table}[H]
    \begin{adjustbox}{width=\columnwidth,center}
    \begin{tabular}{@{}lc>{\scriptsize}rcc>{\scriptsize}rcc>{\scriptsize}rcc>{\scriptsize}r@{}}
    \toprule
    Experiment & 
    \multicolumn{2}{c}{Hologic} & \phantom{ab} &
    \multicolumn{2}{c}{Siemens} & \phantom{ab} &
    \multicolumn{2}{c}{GE} & \phantom{ab} &
    \multicolumn{2}{c}{All} \\
    & AUC & Conf. Int. && AUC & Conf. Int. && AUC & Conf. Int. && AUC & Conf. Int. \\
    \midrule
    Baseline & 0.872 & [0.849, 0.895] && 0.780 & [0.720, 0.841] && 0.863 & [0.811, 0.916] && 0.855 & [0.834, 0.876] \\
    \midrule
    KD & 0.882 & [0.860, 0.904] && 0.806 & [0.748, 0.865] && 0.882 & [0.838, 0.926] && 0.870 & [0.850, 0.889] \\
    SelectiveKD & 0.883 & [0.860, 0.906] && 0.824 & [0.769, 0.879] && 0.892 & [0.846, 0.938] && 0.877 & [0.858, 0.896] \\
    KD* & 0.880 & [0.857, 0.902] && 0.834 & [0.776, 0.891] && 0.862 & [0.808, 0.916] && 0.862 & [0.842, 0.883] \\
    SelectiveKD* & 0.891 & [0.870, 0.912] && 0.862 & [0.814, 0.909] && 0.889 & [0.841, 0.937] && 0.879 & [0.860, 0.897] \\
    \bottomrule
    \end{tabular}
    \end{adjustbox}
    \caption{Main results: 
    the table shows the ROC AUC and the associated 95\% confidence intervals for various ablations of our method. We evaluate on Hologic, Siemens, and GE test splits as well as all 3 together. The asterisk (*) indicates models that incorporated weakly annotated data. We highlight that annotations are only available for cases from the Hologic subset and our approach contributes to especially large performance gains in subsets without annotations (Siemens, GE).}
    \label{tab:main-results}
\end{table}
\vspace{-0.7cm}

\textbf{Efficiency of SelectiveKD} SelectiveKD outperformed the baseline with statistical significance ($p < 0.001$) both with and without using weak annotations, and it also consistently outperformed KD, although the difference was found to be statistically significant (p < 0.001) only when using weak annotations. The best overall performance is achieved with SelectiveKD using weak annotations. We believe there are three main reasons for this; one is that weak annotations also allow the computation of supervised loss for the classification task, adding extra supervision to the model. Second, using weak annotations further mitigates the potential noise introduced by the teacher model by preventing the inclusion of false positive slices from cancer-negative exams, as the inclusion of slices based on model predictions is conditioned on the breast label being positive. Finally, the optimal confidence threshold when using weak annotations was found to be lower than that without using weak annotations (0.1 and 0.7, respectively). This allows the inclusion of more slices in training and makes sense considering that with weak annotations, the model is applied only on cancer-positive exams, while without them, it is applied on both cancer-positive and negative exams; therefore, it requires a higher confidence threshold to filter out false positive slices from cancer-negative exams.

\textbf{Generalization} Interestingly, the biggest improvements in performance were obtained on the Siemens and GE datasets, which can be considered as out-of-domain distribution for the baseline since it is trained only on Hologic data. This shows our method's potential to improve generalization capability. The fact that SelectiveKD consistently outperformed KD on Siemens and GE and that the gap between the two methods is larger for Siemens and GE datasets compared to Hologic provides evidence that the use of PL filtering can mitigate the noise introduced by wrong teacher model predictions, especially when incorporating data that is out of distribution for the teacher model.

\subsection{Annotation cost efficiency}
The previous results show that our semi-supervised framework can be used to leverage unannotated data and reach better performance and generalization. 

In this part, we focus on analyzing how the performance scales with respect to the proportion of annotated data. To that end, we used our annotated Hologic dataset and ran multiple experiments to gradually increase the proportion of annotated exams included in training from 10 to 100\% with 10\% increments. For each 10\% increment, we create a corresponding dataset by randomly sampling 10\% of cancer-positive, benign, and cancer-negative exams, respectively, to ensure the proportion of each subgroup remains constant, which we add to the set of previously sampled annotated exams while the annotations for remaining exams are ignored. In total, 10 different datasets were created that contain the same exams but only the presence of annotation changes.

In this series of experiments, we first trained a baseline model on each dataset to serve as the teacher model for the semi-supervised setting and provide a baseline for comparison. We then applied the SelectiveKD and SelectiveKD with weak annotation settings on each dataset.

\begin{figure}[t]
    \centering
    \includegraphics[width=\columnwidth]{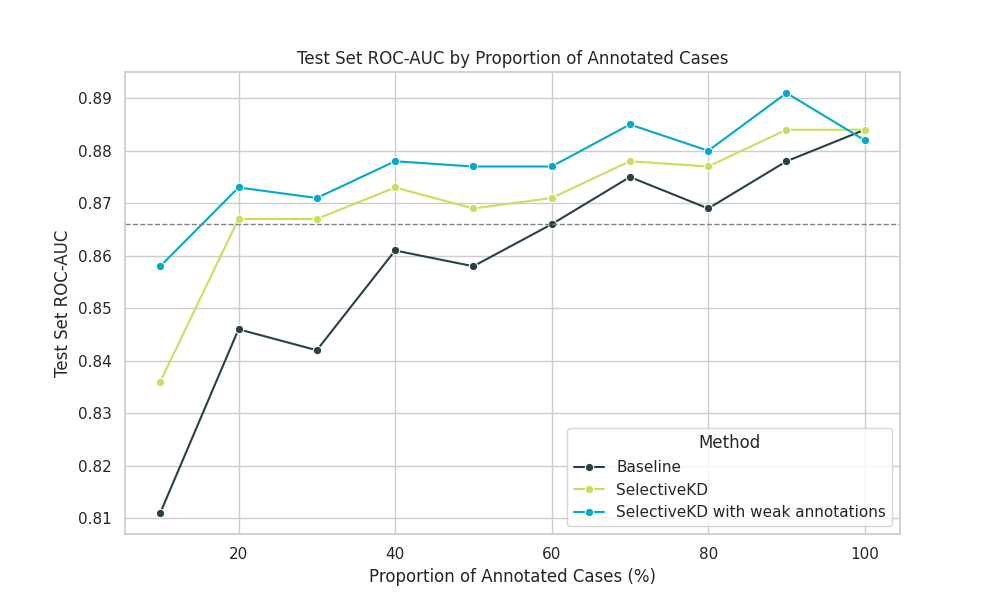}
    \vspace{-0.7cm}
    \caption{
Test set ROC AUC values versus the proportion of annotated exams: The scaling of model performance with respect to the proportion of annotated data is illustrated for the baseline and SelectiveKD. By using our approach, we show that the model achieves similar level of cancer detection performance while using less than one third of annotations (dotted line).
}
    \label{fig:annotation-cost}
\end{figure}

As shown in Fig. \ref{fig:annotation-cost}, our semi-supervised framework can offer meaningful performance improvements, especially when only a small portion of the data is annotated. Using only 20\% of annotated exams, SelectiveKD is able to match the performance of the baseline using 60\% of annotated exams and that of the baseline using 70\% of annotated exams when also using weak annotations. As expected, the gap in performance between all 3 settings decreases as the proportion of annotated exams gets closer to 100\%. Detailed results, including confidence intervals, can be found in the Supplementary Materials.

\vspace{-0.6cm}

\section{Conclusion}
We introduced SelectiveKD, a framework for building a cancer detection CAD for DBT that leverages semi-supervised learning to utilize unannotated slices from DBT stacks. By leveraging knowledge distillation (KD) and selective dataset expansion through pseudo-label (PL) filtering, we showed that we can significantly improve cancer detection performance on a large-scale dataset of over 10,000 DBT exams. Further, we demonstrated that our approach contributes to improving the model's ability to generalize without the need to explicitly collect annotations for each device manufacturer. Finally, our findings suggest that we can build powerful DBT CAD models with fewer annotations by adding more unannotated or weakly annotated cases. 

As limitations of this study, the presented framework mainly focused on improving the classification performance of CAD. We believe a deeper investigation of how our approach impacts lesion localization performance as well will add more clinical value to the model. Further, more thorough investigation of CAD model performance across various patient subgroups across multiple institutions is warranted to build a stronger evidence for the efficacy of the presented framework for real world use. We believe the presented framework and the findings from this study contribute toward increasing the practical value of deep learning based CAD for improving breast cancer screening using DBT.
\newpage
\bibliographystyle{splncs04}
\bibliography{ref}

\begin{thebibliography}{10}
\providecommand{\url}[1]{\texttt{#1}}
\providecommand{\urlprefix}{URL }
\providecommand{\doi}[1]{https://doi.org/#1}

\bibitem{AndhariaDBT}
Andharia, D., Shah, H., Prajapati, A.D., Bhansali, A.D., Shah, A., Desai, D.: Digital breast tomosynthesis(dbt) vs 2d mammography and impact of combined use: A meta-analysis. In: medRxiv (2023)

\bibitem{bahl2018breast}
Bahl, M., Gaffney, S., McCarthy, A.M., Lowry, K.P., Dang, P.A., Lehman, C.D.: Breast cancer characteristics associated with 2d digital mammography versus digital breast tomosynthesis for screening-detected and interval cancers. Radiology  \textbf{287}(1),  49--57 (2018)

\bibitem{berthelot2019mixmatch}
Berthelot, D., Carlini, N., Goodfellow, I., Papernot, N., Oliver, A., Raffel, C.A.: Mixmatch: A holistic approach to semi-supervised learning. Advances in neural information processing systems  \textbf{32} (2019)

\bibitem{chen2022semi}
Chen, Y., Mancini, M., Zhu, X., Akata, Z.: Semi-supervised and unsupervised deep visual learning: A survey. IEEE transactions on pattern analysis and machine intelligence  (2022)

\bibitem{chen2020semi}
Chen, Y., Zhu, X., Li, W., Gong, S.: Semi-supervised learning under class distribution mismatch. In: Proceedings of the AAAI Conference on Artificial Intelligence. vol.~34, pp. 3569--3576 (2020)

\bibitem{chikarmane2021screening}
Chikarmane, S.A., Cochon, L.R., Khorasani, R., Sahu, S., Giess, C.S.: Screening mammography performance metrics of 2d digital mammography versus digital breast tomosynthesis in women with a personal history of breast cancer. American Journal of Roentgenology  \textbf{217},  587--594 (2021)

\bibitem{delong1988comparing}
DeLong, E.R., DeLong, D.M., Clarke-Pearson, D.L.: Comparing the areas under two or more correlated receiver operating characteristic curves: a nonparametric approach. Biometrics pp. 837--845 (1988)

\bibitem{DhamijaDBT}
Dhamija, E., Gulati, M., Deo, S., Gogia, A., Hari, S.: Digital breast tomosynthesis: an overview. Indian Journal of Surgical Oncology  \textbf{12}(2),  315--329 (2021)

\bibitem{goldberg2022new}
Goldberg, J.E., Reig, B., Lewin, A.A., Gao, Y., Heacock, L., Heller, S.L., Moy, L.: New horizons: artificial intelligence for digital breast tomosynthesis. RadioGraphics  \textbf{43}(1),  e220060 (2022)

\bibitem{hinton2015distilling}
Hinton, G., Vinyals, O., Dean, J.: Distilling the knowledge in a neural network (2015)

\bibitem{Huang2023SpectralSpatialMT}
Huang, L., Chen, Y., He, X.: Spectral–spatial masked transformer with supervised and contrastive learning for hyperspectral image classification. IEEE Transactions on Geoscience and Remote Sensing  \textbf{61},  1--18 (2023)

\bibitem{ko2021accuracy}
Ko, M.J., Park, D.A., Kim, S.H., Ko, E.S., Shin, K.H., Lim, W., Kwak, B.S., Chang, J.M.: Accuracy of digital breast tomosynthesis for detecting breast cancer in the diagnostic setting: a systematic review and meta-analysis. Korean journal of radiology  \textbf{22}(8), ~1240 (2021)

\bibitem{lee2013pseudo}
Lee, D.H., et~al.: Pseudo-label: The simple and efficient semi-supervised learning method for deep neural networks. In: Workshop on challenges in representation learning, ICML. vol.~3, p.~896. Atlanta (2013)

\bibitem{lee2023transformer}
Lee, W., Lee, H., Lee, H., Park, E.K., Nam, H., Kooi, T.: Transformer-based deep neural network for breast cancer classification on digital breast tomosynthesis images. Radiology: Artificial Intelligence  \textbf{5}(3),  e220159 (2023)

\bibitem{loshchilov2016sgdr}
Loshchilov, I., Hutter, F.: Sgdr: Stochastic gradient descent with warm restarts. arXiv preprint arXiv:1608.03983  (2016)

\bibitem{Nguyendbt}
Nguyen, T., Levy, G., Poncelet, E., Le~Thanh, T., Prolongeau, J., Phalippou, J., Massoni, F., Laurent, N.: Overview of digital breast tomosynthesis: Clinical cases, benefits, and disadvantages. Diagnostic and interventional imaging  \textbf{96}(9),  843--859 (2015)

\bibitem{Shahan2016AnOO}
Shahan, C.L.: An overview of digital breast tomosynthesis. The West Virginia medical journal  \textbf{2016}, ~996 (2016)

\bibitem{Shen2023FullyCS}
Shen, Y., Shi, L., Zhao, J., Dong, Y., Wang, L.: Fully convolutional spectral–spatial fusion network integrating supervised contrastive learning for hyperspectral image classification. IEEE Journal of Selected Topics in Applied Earth Observations and Remote Sensing  \textbf{16},  9077--9088 (2023)

\bibitem{shoshan2022artificial}
Shoshan, Y., Bakalo, R., Gilboa-Solomon, F., Ratner, V., Barkan, E., Ozery-Flato, M., Amit, M., Khapun, D., Ambinder, E.B., Oluyemi, E.T., et~al.: Artificial intelligence for reducing workload in breast cancer screening with digital breast tomosynthesis. Radiology  \textbf{303}(1),  69--77 (2022)

\bibitem{singh2023shifting}
Singh, P., Chukkapalli, R., Chaudhari, S., Chen, L., Chen, M., Pan, J., Smuda, C., Cirrone, J.: Shifting to machine supervision: Annotation-efficient semi and self-supervised learning for automatic medical image segmentation and classification. arXiv preprint arXiv:2311.10319  (2023)

\bibitem{tamCXR}
Tam{\'e}, I.d.A., Sirotkin, K., Carballeira, P., Escudero-Vi{\~n}olo, M.: Self-supervised curricular deep learning for chest x-ray image classification. arXiv preprint arXiv:2301.10687  (2023)

\bibitem{tardy2021trainable}
Tardy, M., Mateus, D.: Trainable summarization to improve breast tomosynthesis classification. In: International Conference on Medical Image Computing and Computer-Assisted Intervention. pp. 140--149. Springer (2021)

\bibitem{xie2020self}
Xie, Q., Luong, M.T., Hovy, E., Le, Q.V.: Self-training with noisy student improves imagenet classification. In: Proceedings of the IEEE/CVF conference on computer vision and pattern recognition. pp. 10687--10698 (2020)

\bibitem{zhang2022semisupervised}
Zhang, J., Yang, J., Yu, J., Fan, J.: Semisupervised image classification by mutual learning of multiple self-supervised models. International Journal of Intelligent Systems  \textbf{37}(5),  3117--3141 (2022)

\end{thebibliography}

\end{document}


\title{Supplementary Materials: SelectiveKD: A semi-supervised framework for cancer detection in DBT through Knowledge Distillation and Pseudo-labeling}
%
%
\author{Laurent Dillard\inst{1}\thanks{These authors contributed equally to this work.}, Hyeonsoo Lee\inst{1,*}, Weonsuk Lee\inst{1,*}, Tae Soo Kim\inst{1}, Ali Diba\inst{1} and Thijs Kooi\inst{1}}
%

\institute{
Lunit Inc, Seoul, Republic of Korea\\
\email{\{laurent.dillard, hslee, iwonseok5762, taesoo.kim, ali, tkooi\}@lunit.io}}
%
\maketitle


\begin{table}[htbp]
\centering
\begin{adjustbox}{center}
\begin{tabular}{lc|c|c|c|c}
\toprule
& \textbf{Normal} & \textbf{Fu-Benign} & \textbf{Bx-Benign} & \textbf{Cancer} & \textbf{Total} \\
\midrule
\textbf{Training} \\
\midrule
Hologic & 1,210 & 1,361 & 1,143 & 1,025 & 4,739 \\
Siemens & 1,037 & 1,056 & 421 & 283 & 2,797 \\
GE & 157 & 1,081 & 215 & 185 & 1,638 \\
\textbf{Total} & 2,404 & 3,498 & 1,779 & 1,493 & 9,174 \\
\midrule
\textbf{Validation} \\
\midrule
Hologic & 248 & 159 & 282 & 332 & 1,021 \\
Siemens & 213 & 155 & 129 & 103 & 600 \\
GE & 36 & 183 & 86 & 62 & 367 \\
\textbf{Total} & 497 & 497 & 497 & 497 & 1,988 \\
\midrule
\textbf{Test} \\
\midrule
Hologic & 252 & 190 & 284 & 337 & 1,063 \\
Siemens & 212 & 139 & 142 & 84 & 577 \\
GE & 35 & 168 & 71 & 76 & 350 \\
\textbf{Total} & 497 & 497 & 497 & 497 & 1,988 \\
\bottomrule
\end{tabular}
\end{adjustbox}
\label{tab:data_stats}
\caption{Comprehensive Data Distribution for Training, Validation, and Test Sets: The table details the number of cases across different diagnostic categories including normal, follow-up benign (Fu-Benign), biopsy-proven benign (Bx-Benign), and cancer for each manufacturer—Hologic, Siemens, and GE. Totals are provided for each category within the training, validation, and test sets to give a clear overview of the dataset composition.}
\end{table}


\begin{table}[H]
    \begin{adjustbox}{width=\columnwidth,center}
    \begin{tabular}{@{}lc>{\scriptsize}rcc>{\scriptsize}rcc>{\scriptsize}rcc>{\scriptsize}r@{}}
    \toprule
    Threshold & 
    \multicolumn{2}{c}{Hologic} & \phantom{ab} &
    \multicolumn{2}{c}{Siemens} & \phantom{ab} &
    \multicolumn{2}{c}{GE} & \phantom{ab} &
    \multicolumn{2}{c}{All} \\
    & AUC & Conf. Int. && AUC & Conf. Int. && AUC & Conf. Int. && AUC & Conf. Int. \\
\midrule
0.1 & 0.8911 & [0.8699 - 0.9122] && 0.8618 & [0.8143 - 0.9093] && 0.8892 & [0.8410 - 0.9374] && 0.8786 & [0.8601 - 0.8970] \\
\midrule
0.2 & 0.8850 & [0.8630, 0.9070] && 0.8480 & [0.7940, 0.9020] && 0.8840 &  [0.8370, 0.9310] && 0.8650 & [0.8450, 0.8850] \\
\midrule
0.3 & 0.8800 & [0.8570, 0.9040] && 0.8510 & [0.8000, 0.9030] && 0.9030 & [0.8630, 0.9430] && 0.8660 & [0.8470, 0.8860] \\
\midrule
0.4 & 0.8740 & [0.8500, 0.8980] && 0.8240 & [0.7640, 0.8850] && 0.8690 & [0.8190, 0.9180] && 0.8640 & [0.8430, 0.8840] \\
\midrule
0.5 & 0.8880 & [0.8660, 0.9100] && 0.8310 & [0.7730, 0.8890] && 0.8880 & [0.8430, 0.9330] && 0.8710 & [0.8510, 0.8900] \\
\midrule
0.7 & 0.8850 & [0.8620, 0.9070] && 0.8290 & [0.7700, 0.8870] && 0.8680 & [0.8150, 0.9200] && 0.8640 & [0.8430, 0.8840] \\
\midrule
0.9 & 0.8650 & [0.8410, 0.8900] && 0.8120 & [0.7520, 0.8720] && 0.8580 & [0.8050, 0.9110] && 0.8430 & [0.8220, 0.8650] \\
    \bottomrule
    \end{tabular}
    \end{adjustbox}
    \caption{A pseudo labeling threshold grid search result of SelectiveKD when weak annotations are available.}
    \label{tab:main-results}
\end{table}
\vspace{-0.7cm}


\begin{table}[H]
    \begin{adjustbox}{width=\columnwidth,center}
    \begin{tabular}{@{}lc>{\scriptsize}rcc>{\scriptsize}rcc>{\scriptsize}rcc>{\scriptsize}r@{}}
    \toprule
    Threshold & 
    \multicolumn{2}{c}{Hologic} & \phantom{ab} &
    \multicolumn{2}{c}{Siemens} & \phantom{ab} &
    \multicolumn{2}{c}{GE} & \phantom{ab} &
    \multicolumn{2}{c}{All} \\
    & AUC & Conf. Int. && AUC & Conf. Int. && AUC & Conf. Int. && AUC & Conf. Int. \\
\midrule
0.1 & 0.8870 & [0.8650, 0.9090] && 0.8050 & [0.7480, 0.8610] && 0.8740 & [0.8250, 0.9240] && 0.8680 & [0.8480, 0.8870] \\
\midrule
0.2 & 0.8790 & [0.8560, 0.9010] && 0.8230 & [0.7670, 0.8800] && 0.8730 & [0.8260, 0.9200] && 0.8680 & [0.8480, 0.8870] \\
\midrule
0.3 & 0.8880 & [0.8670, 0.9100] && 0.8200 & [0.7660, 0.8750] && 0.8790 & [0.8330, 0.9250] && 0.8750 & [0.8560, 0.8940] \\
\midrule
0.4 & 0.8850 & [0.8620, 0.9080] && 0.8140 & [0.7590, 0.8690] && 0.8990 & [0.8580, 0.9400] && 0.8760 & [0.8570, 0.8950] \\
\midrule
0.5 & 0.8820 & [0.8590, 0.9050] && 0.8180 & [0.7650, 0.8710] && 0.8860 & [0.8420, 0.9300] && 0.8740 & [0.8550, 0.8930] \\
\midrule
0.7 & 0.8833 & [0.8604 - 0.9063] && 0.8240 & [0.7689 - 0.8791] && 0.8919 & [0.8459 - 0.9379] && 0.8769 & [0.8576 - 0.8961] \\
\midrule
0.9 & 0.8670 & [0.8420, 0.8910] && 0.8240 & [0.7670, 0.8810] && 0.8610 & [0.8110, 0.9120] && 0.8640 & [0.8440, 0.8840] \\
    \bottomrule
    \end{tabular}
    \end{adjustbox}
    \caption{A pseudo labeling threshold grid search result of SelectiveKD when weak annotations are NOT available.}
    \label{tab:main-results}
\end{table}
\vspace{-0.7cm}


\begin{table}[H]
    \centering
    \begin{adjustbox}{width=\columnwidth,center}
    \begin{tabular}{ccccccccccc}
    \toprule
         Experiment & 10\% & 20\% & 30\% & 40\% & 50\% & 60\% & 70\% & 80\% & 90\% & 100\%\\
    \midrule
         Baseline & 0.811 & 0.846 & 0.842 & 0.861 & 0.858 & 0.866 & 0.875 & 0.869 & 0.878 & 0.884 \\
         95\% CI & \scriptsize{[0.782, 0.839]} & \scriptsize{[0.821, 0.872]} & \scriptsize{[0.817, 0.868]} & \scriptsize{[0.0.836, 0.0.885]} & \scriptsize{[0.833, 0.883]} & \scriptsize{[0.843, 0.890]} & \scriptsize{[0.852, 0.898]} & \scriptsize{[0.846, 0.892]} & \scriptsize{[0.856, 0.900]} & \scriptsize{[0.862, 0.906]}\\
    \midrule
         SelectiveKD & 0.836 & 0.867 & 867 & 0.873 & 0.869 & 0.871 & 0.879 & 0.877 & 0.884 & 0.884 \\
         95\% CI & \scriptsize{[0.809, 0.863]} & \scriptsize{[0.843, 0.891]} & \scriptsize{[0.843, 0.891]} & \scriptsize{[0.850, 0.897]} & \scriptsize{[0.845, 0.893]} & \scriptsize{[0.848, 0.895]} & \scriptsize{[0.856, 0.902]} & \scriptsize{[0.855,0.900]} & \scriptsize{[0.861, 0.906]} & \scriptsize{[0.862, 0.905]}\\
    \midrule
         SelectiveKD* & 0.858 & 0.873 & 0.871 & 0.878 & 0.877 & 0.877 & 0.885 & 0.880 & 0.891 & 0.882 \\
         95\% CI & \scriptsize{[0.833, 0.883]} & \scriptsize{[0.850, 0.897]}  &  \scriptsize{[0.848, 0.895]} & \scriptsize{[0.855, 0.900]}  & \scriptsize{[0.854, 0.900]} 
 & \scriptsize{[0.854, 0.900]}  & \scriptsize{[0.864, 0.906]} & \scriptsize{[0.857, 0.902]} & \scriptsize{[0.869, 0.912]}  & \scriptsize{[0.860, 0.904]} \\
    \bottomrule
    \end{tabular}
    \end{adjustbox}
    \caption{Annotation cost efficiency results. * denotes use of weakly annotated data. Comparison of Baseline, SelectiveKD and SelectiveKD with weakly annotated data when gradually increasing the proportion of annotated exams in the training data. Performance metric is ROC-AUC reported along with 95\% confidence intervals.}
    \label{tab:supp_annotation_cost}
\end{table}